\documentclass[12pt]{iopart}

\usepackage{iopams}  

\expandafter\let\csname equation*\endcsname\relax
\expandafter\let\csname endequation*\endcsname\relax
\usepackage{amsmath}

\usepackage{cite}  

\usepackage{tikz}

\usepackage{amsfonts}
\usepackage{bbm}
\DeclareMathAlphabet{\mathbbold}{U}{bbold}{m}{n}
\usepackage{algorithm}      
\usepackage[noend]{algpseudocode}   
\algrenewcommand\alglinenumber[1]{\scriptsize #1:}

\usepackage{listings}       

\usepackage{bbold}

\usepackage{url}

\usepackage{subfig}
\usepackage{graphicx}

\usepackage{hyperref}
\usepackage{cleveref}
\usepackage{comment}

\usepackage{orcidlink}

\newcommand{\enc}{\mathrm{Enc}}
\newcommand{\dec}{\mathrm{Dec}}

\newcommand{\ikmeans}{\mathrm{IKMeans}}

\usepackage{pifont}
\definecolor{fiorentina}{RGB}{72, 46, 146}

\RequirePackage{xspace}
\makeatletter
\DeclareRobustCommand\onedot{\futurelet\@let@token\@onedot}
\def\@onedot{\ifx\@let@token.\else.\null\fi\xspace}

\def\eg{\emph{e.g}\onedot} 

\def\ie{\emph{i.e}\onedot}

\begin{document}

\def\aligned{\vcenter\bgroup\let\\\cr
\halign\bgroup&\hfil${}##{}$&${}##{}$\hfil\cr}
\def\endaligned{\crcr\egroup\egroup}

\title[Towards virtual painting recolouring using Vision Transformer on XRF]{Towards virtual painting recolouring using Vision Transformer on X-Ray Fluorescence datacubes}

\author{Alessandro Bombini\orcidlink{0000-0001-7225-3355}${}^{1,2}$, Fernando García-Avello Bofías\orcidlink{0000-0001-6640-8736}${}^{1}$, Francesca Giambi${}^{3}$\orcidlink{0000-0001-6446-7185},  Chiara Ruberto\orcidlink{0000-0003-0321-7160}${}^{1,4}$.
}

\address{${}^{1}$ Istituto Nazionale di Fisica Nucleare (INFN), Via Bruno Rossi 1, 50019 Sesto Fiorentino (FI), Italy}
\address{${}^{2}$ ICSC - Centro Nazionale di Ricerca in High Performance Computing, Big Data \& Quantum Computing, Via Magnanelli 2, 40033 Casalecchio di Reno (BO), Italy}
\address{${}^{3}$ Università degli Studi di Firenze, P.za di San Marco 4, 50121 Firenze (FI), Italy}
\address{${}^{4}$ Dipartimento di Fisica, Università degli Studi di Firenze, Via Giovanni Sansone 1, 50019 Sesto Fiorentino (FI), Italy}

\ead{bombini@fi.infn.it}

\vspace{10pt}
\begin{indented}
\item[]\today
\end{indented}

\begin{abstract}
In this contribution, we define (and test) a pipeline to perform virtual painting recolouring using raw data of X-Ray Fluorescence (XRF) analysis on pictorial artworks. To circumvent the small dataset size, we generate a synthetic dataset, starting from a database of XRF spectra; furthermore, to ensure a better generalisation capacity (and to tackle the issue of in-memory size and inference time), we define a Deep Variational Embedding network to embed the XRF spectra into a lower dimensional, K-Means friendly, metric space. 
  We thus train a set of models to assign coloured images to embedded XRF images. We report here the devised pipeline performances in terms of visual quality metrics, and we close on a discussion on the results.
\end{abstract}

%
\vspace{2pc}
\noindent{\it Keywords}: Computer Vision and Cultural Heritage $\cdot$ Vision Transformers $\cdot$ MA-XRF $\cdot$ Heritage Science $\cdot$ Computer Vision
%
%
%
%

\section{Introduction}
\label{sec:intro}

The rise of Artificial Intelligence in the last decade had a tremendous impact on all academic fields. Statistical and deep learning methods have been ubiquitously employed to help researchers in their tasks, from the easier ones to the most complex ones. Yet, in many fields, there is still room to find suitable applications of such technologies. 

In the field of nuclear physics for Cultural Heritage analysis and imaging (for a non-exhaustive list of relevant reviews to the subject, see, e.g.~\cite{alma991021183529704336, KnollBookXRF, MandoPixe, Grieken1993HandbookOX, jenkins1995quantitative, Janssens2000MicroscopicXF, Verma2007AtomicAN, Ruberto2023} and references therein), the use of deep learning methods has begun and is spreading; for example, deep learning can be used to help researcher analyse data from non-destructive testing techniques, such as neutron imaging, gamma-ray spectroscopy, and/or X-ray based imaging techniques. These algorithms can help to identify patterns and correlations in the
data, providing valuable insights into the artwork’s composition, age, and condition \cite{Kleynhans2020,5967899,8432512,8667664,666999xrfdl, D2JA00114D, Jones2022, arxiv.2207.12651, Dirks2022AutoEncoderNN, Liu2023NeuralNF} (for other Machine Learning approaches in Cultural Heritage, see \cite{FIORUCCI2020102}, and references therein). 

A relevant technique in the field of nuclear technologies applied to Cultural Heritage is X-Ray Fluorescence (XRF) spectrometry, especially in its Macro mapping version (MA-XRF).  The XRF technique is employed in Cultural Heritage applications, since no sample pretreatment is required. Furthermore, XRF allows for non-invasive, non-destructive measurements, which is crucial in maintaining the integrity of the sample and the repeatably of the measurements (see \cite{KnollBookXRF,MandoPixe}, and references therein). 

In Macro X-Ray Fluorescence mapping (MA-XRF), the imaging apparatus produces a data cube of a scanned area of an artwork; each datacube pixel is formed by a spectrum containing fluorescence lines associated with the element composition of the pigment present in the pictorial layers. From there, it is possible to extract elemental distribution maps of the scanned area of the artwork, proving to be extremely useful for material characterisation, and also for understanding an artist’s painting and production techniques \cite{alfeld2020ma}. 

One possible application of Computer Vision techniques on data obtained form such techniques is \textit{virtual colouring}, \ie the extraction of a RGB image out of a MA-XRF data cube of a pictorial artwork analysis. This task may be relevant in conservation science, since techniques such as MA-XRF are capable of detecting spectral signal (thus containing elemental information on pictorial pigments) even from degraded surfaces (either paintings or frescoes) \cite{C6JA00249H, Fiorillo2020, Ruberto2023}, and also to study a series of subsurface features such as over-painted compositions, pentimenti, covered paint losses, retouching and other physical changes\cite{Dik2008, dredge2015mapping, martins2016jackson, VANDERSNICKT2018238}. Thus, such pipeline may assist conservation scientist and heritage science experts in the interpretation of the MA-XRF data, granting an additional visual feedback (\ie, the reconstructed RGB image).

Unfortunately, the availability of quality dataset for many of the learning task is usually hampered by various causes, such as slow data creation due to intrinsic apparatus speed limits, legal/Intellectual Property strict limits, difficulties in accessing study objects, limited availability and high cost of experimental apparatus, the necessity of high skilled and qualified personnel to use it, etc. Furthermore, data standardisation is far from the usual in other fields, due to the different, custom nature of the various measuring apparatus, as well as different measurement conditions, different detectors, etc. 

To tackle this issue, we define a pipeline to be applied use case by use case, capable of working without huge dataset. To do so, we resort to synthetic MA-XRF data generation, starting from a tabulated database of pigments' XRF signal, and Deep Variational Embedding, to (a) reduce the disk size of the generated synthetic dataset, and (b) to extract relevant features and reduce the role of different measuring apparatus, conditions, and environmental impacts. Afterwards, we train a Vision Transformer to map synthetic, embedded MA-XRF maps to RGB images. 

This study is the first step towards the applicability of such pipeline in real-case scenario, where a domain adaptation learning technique will be added to apply, in an almost self-supervising manner, the whole pipeline to a real MA-XRF image obtained from an analysis conducted on a use case. 

In \cref{sec:methods} we describe the building blocks of the aforementioned virtual recolouring pipeline; in \cref{sec:results} we report the results of the training,  and the inference of the pipeline on the test set. We close with \cref{sec:concl} with a discussion. Finally, in \cref{sec:data-and-code} we report the link to the git repository containing the code used to produce the results presented in this contribution.

\section{Methods: the virtual colouring pipeline}
\label{sec:methods}

\begin{figure}
    \centering
    \includegraphics[width=0.98\linewidth]{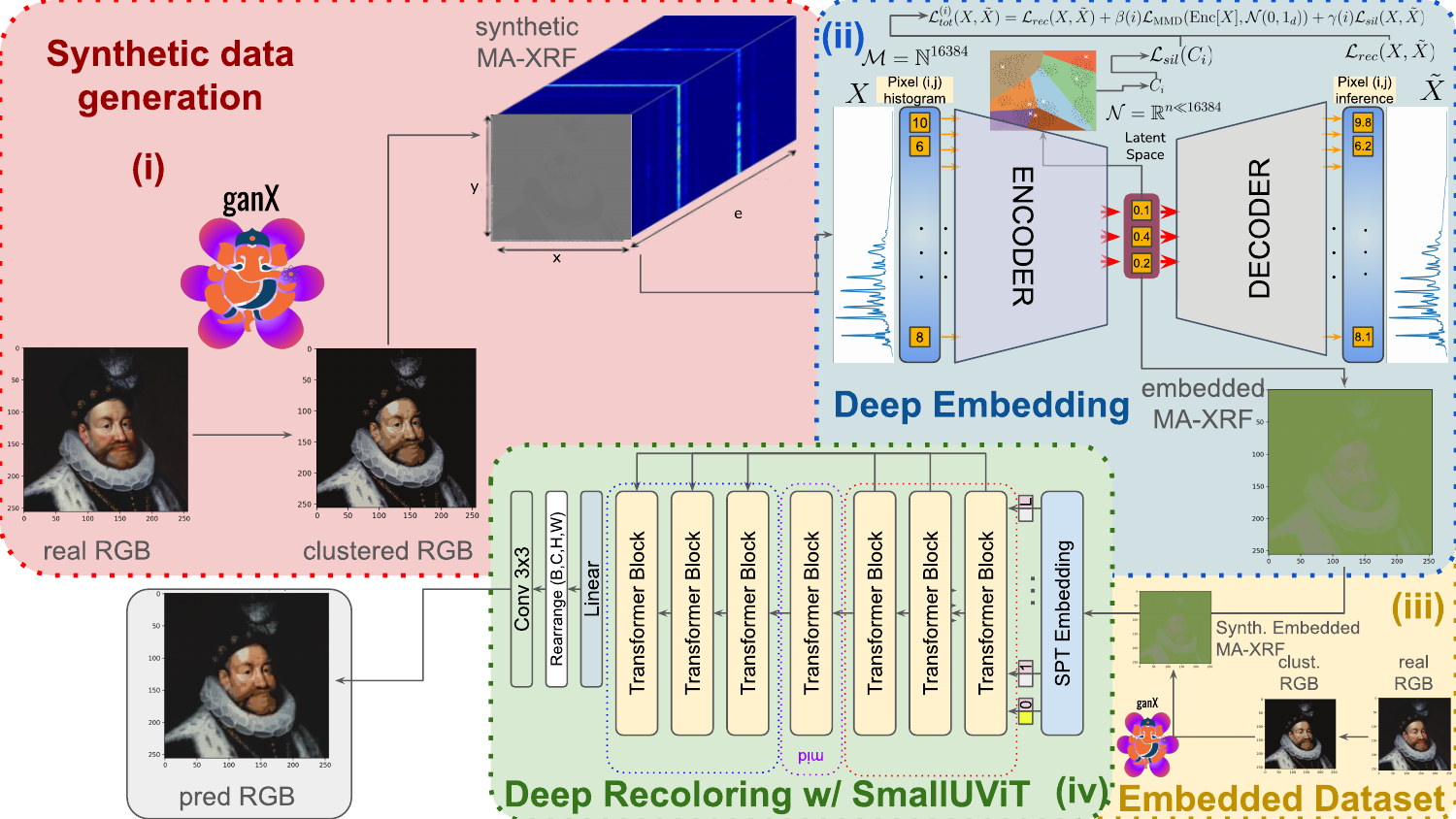}
    \caption{Visual Abstract of the pipeline devised in this Paper. The flow is described in the main text, and it comprises 4 steps: Synthetic Data Generation of spectral signal (in the red region in the figure); a trained Deep Embedding model (in the blue region); an embedded synthetic dataset of embedded MA-XRF images (in the yellow region), and finally a computer vision model to perform virtual recoloring. }
    \label{fig:visual_abstract}
\end{figure}

In this section we describe the pipeline architecture design and implementation. The goal of the whole process is to assign a coloured image to a MA-XRF map, obtained using the movable apparatus developed at the INFN-CHNet node facility in Florence, LABEC (Laboratorio di tecniche nucleari per l'Ambiente e i BEni Culturali)\cite{Taccetti2019,app11083462,Chiari2021,Ruberto2023}.

The pipeline relies on the following steps:
\begin{enumerate}
    \item Generate a synthetic dataset of Spectral signal;
    \item train a Deep Embedding model to map spectra into lower dimensional metric space;
    \item Starting from RGB images, build a synthetic dataset of embedded MA-XRF images using the python package \textsc{ganX} \cite{bombini2023ganx}\footnote{Either by creating full-fledged MA-XRF images and then embed them in the low dimensional latent space using the Deep Embedding model, or generate them directly in latent space.};
    \item Train a Computer Vision model to assign RGB images to (embedded) MA-XRF images. 
\end{enumerate}
This pipeline is visually reported in \cref{fig:visual_abstract}.

\subsection{Motivating the pipeline architecture}
X-ray Fluorescence (XRF) is a non-invasive, non-destructive analytical technique widely used in the study of cultural heritage, and it is well suited for analysis of pictorial artworks (for a nice introduction to the subject, see, \eg~\cite{alma991021183529704336, KnollBookXRF, MandoPixe, Grieken1993HandbookOX, jenkins1995quantitative, Janssens2000MicroscopicXF, Verma2007AtomicAN, Ruberto2023} and references therein).

XRF employs X-ray beams emitted by radiogenic tube to excite atoms within the material composing the study object. As a result of the matter-radiation interaction, these atoms emit characteristic X-ray fluorescence radiation, which is collected in appropriate detectors, defining a spectrogram (\ie, a histogram of counts in the X-ray range); by looking as such spectrogram, researchers are able to identify the chemical elements present in the material. 

Due to material inhomogeneities (always present in the CH study objects), standard single spot XRF analysis may produce false results concerning the chemical composition of the substance under investigation. Thus, it has been developed the XRF macro-mapping imaging technique, known as Macro X-Ray Fluorescence mapping (MA-XRF), with scanning mode acquisition systems. MA-XRF allows to gather data on the sample's material composition and the distribution of the distinctive components within the scanned area, which allows for the creation of elemental distribution maps \cite{alfeld2020ma}. 

The MA-XRF raw data can then be arranged into \textit{spectral datacubes}, \ie a 3-D array $I_{x,y;e}$; the first two indeces are the (discretised) pixel positions, while the third index is the discretised energy (in KeV)/wavelength of the emitted fluorescence radiation. These datacubes may be of different sizes, depending on the $X-Y$ motors range used in the analysis set up, but usually have a fixed Energy depth, due to the Analog-to-Digital converter (ADC) used\footnote{For the apparatus employed in this work, the ADC has a 14-bit memory, thus having spectra with $2^{14}=16384$ energy bins.}. Nevertheless, to maintain spectra readability, the number of energy bins cannot be too low, \ie, no less that $O(500)$ bins; this means that \textit{each} MA-XRF image\footnote{Here and in the following we use MA-XRF image and MA-XRF datacube as synonyms. The images obtained from the datacube by integrating the spectral signal in a certain energy interval are called \textit{elemental maps}.} has a large disk occupation size (approximately, each MA-XRF image, once rebinned to, let us say, 512 bins, has a size which is $512/3\sim 170$ times \textit{bigger} than an RGB image of the same height and width).

Furthermore, due to instrumental limitations, copyright and data management issues, as well as study objects availability, it is almost impossible to have a dataset of standardised MA-XRF images whose size is really suited for Deep Learning applications. 

Additionally, the obtained spectra posses a set of features which are not related to the analysed object, but on the apparatus \textit{in se} (for more information about the physics of the XRF, see \cite{Grieken1993HandbookOX, jenkins1995quantitative, Janssens2000MicroscopicXF, Verma2007AtomicAN}; while, for a more detailed description of difficulties experienced while applying learning techniques on imaging data of Cultural Heritage object, see
\cite{Kleynhans2020,5967899,8432512,8667664,666999xrfdl, D2JA00114D, Jones2022, arxiv.2207.12651, BombiniICIAP2021, BombiniICCSA2022, bombini2023ganx, bombini2024datacube, BombiniThespianXRF, Dirks2022AutoEncoderNN, Liu2023NeuralNF, Ruberto2023, zheng2024deep, Velibor2024deep}, and references therein).

To address all these issue, we resort to
\begin{itemize}
    \item Synthetic dataset generation;
    \item Deep Embedding in a low dimensional metric feature space;
\end{itemize}
The deep embedding is useful either to extract relevant, apparatus-independent features of the material/pigment composition of the target object, and to reduce the in-memory size of MA-XRF images. 

Furthermore, it is possible to add an \textit{unsupervised}, \textit{domain adaptation} \cite{ganin2016domainadversarial} step in the Deep Embedding model training, which may help in the inference of the whole recolouring pipeline on \textit{real} MA-XRF images.

\subsection{Preparing the synthetic dataset - part I: spectra} \label{subsec:datasetI}

To generate a meaningful Spectral Dataset, we started from a database of pigments' XRF signal \cite{infraart10.1145/3593427}. We extracted from there a subset of relevant pigments (a \textit{palette}), comprising different pigments (Red pigments: Red Ochre, Cinnabar; Blue pigments: Cobalt Blue, Smaltino; Yellow pigments: Gold Ochre,  Dark Ochre; Green pigments: Aegirine,  Green Earth; Dark pigments: Caput Mortum, Ivory Black, Carbon Black; White pigment: Titanium white). This palette have been chosen due to its similarity with the palette used to create a mock-up, prepared in our laboratory: a replica of an Etruscan
wall painting from IV century BC from the “Tomba della Quadriga Infernale”\footnote{\url{http://www.museoetruscosarteano.it/pagina4.php}.} in Sarteano (Siena, IT), obtained following the historical buon fresco technique \cite{DELVESCOVO20051015}. 

From the aforementioned palette of couples [pigment XRF signal, pigment RGB], we have built a synthetic dataset of approximately $312\,000$ images (we split them into 70\% training set, 20\% validation set, and 10\% test set) using the Python package \textsc{ganX} \cite{bombini2023ganx}.

As a seed, we have used all the images of Paintings and Frescoes available on wikidata \cite{wikidata_paper}, and freely downloadable from there through the wikidata SPARQL query service\footnote{\url{https://query.wikidata.org/}.}.

The MA-XRF generation system works as in \cite{bombini2023ganx};  starting from the aforementioned palette, the data generation algorithm comprises two main parts:
\begin{enumerate}
    \item A RGB clustering of the input image(s), performed with an iterative K-Means ($\ikmeans$) algorithm, to reduce the colour levels;
    \item A random extraction (Monte Carlo method) to generate a XRF spectra for each pixel, after having unsupervisedly associated a (set of) pigment(s) to each RGB pixel.
\end{enumerate}

To perform step 2., the algorithm computes, pixel-by-pixel, a similarity measure in RGB space (based on the Delta CIEDE 2000 distance \cite{ciede2000paper,ciede2000implementation}), between the (clustered) RGB and the palette indexed pigments' characteristic RGB. Furthermore, to (unsupervisedly) reduce the number of pigments for each pixel, an hard thresholding is performed on the similarity values (values below the threshold are set to zero). This process defines, for each pixel, a linear combination of XRF signals, seen as a probability distribution defining the probability that a certain signal is detected by the detector.

The full pseudocode of the MA-XRF data generation is reported in \cref{alg:GenerateMA-XRF} (from \cite{bombini2023ganx}).
\begin{algorithm}[H]
\caption{Generate MA-XRF}\label{alg:GenerateMA-XRF}
\scriptsize
\begin{algorithmic}[1]
\Require \textsc{RGBclustering}, \textsc{distrToXrf}, \textsc{colorSimilarity} \Comment{Import relevant methods}

\Procedure{generateXRF}{pigments\_dict\_data = $\mathcal{D}_{{color}, {hist}}$, rgb\_img = $R$}
\State $R \leftarrow$ \textsc{RGBclustering}($R$)
\State init $X$; \Comment{Initialise empy XRF data cube}
\For{$r\in R$} \Comment{Iterate over clusters}
    \State init $d$; \Comment{Initialise empty distribution}
    \For{$(c, h) \in \mathcal{D}$} \Comment{Iterate over database}
        \State $\alpha \leftarrow$  \textsc{colorSimilarity}($c, r$) 
        \If{$\alpha \ge \alpha_{th}$} 
            \State $d \leftarrow d + \alpha \cdot h$          \Comment{Sum thresholded distribution}   
        \EndIf
    \EndFor

    \State $d \leftarrow d /||{d}||$ \Comment{Normalise distribution}
    \State $X[\textrm{idx}(X)== r] \leftarrow $ \textsc{distrToXrf}($d$) \Comment{Use Monte Carlo method only on the Cluster's pixel}
\EndFor 

\State \textbf{return} $X$

\EndProcedure
\end{algorithmic}
\end{algorithm}

\subsubsection{The spectra dataset}\phantom{.}

From the dataset of MA-XRF images we randomly extracted a subset of 5 million spectra (by selecting random pixels spectra). We divided this set with a 70/20/10 \% split for Training, Validation and Test, so that we have 
\begin{itemize}
    \item Train set shape: \hspace{0.68cm} $[3.5\cdot 10^6, 512]$
    \item Validation set shape: $[1.0\cdot 10^6, 512]$
    \item Test set shape: \hspace{0.89cm} $[0.5\cdot 10^6, 512]$
\end{itemize}

\subsection{Training the Deep Embedding Model}\label{subsec:ae}

\begin{figure}
    \centering
    \includegraphics[width=0.95\linewidth]{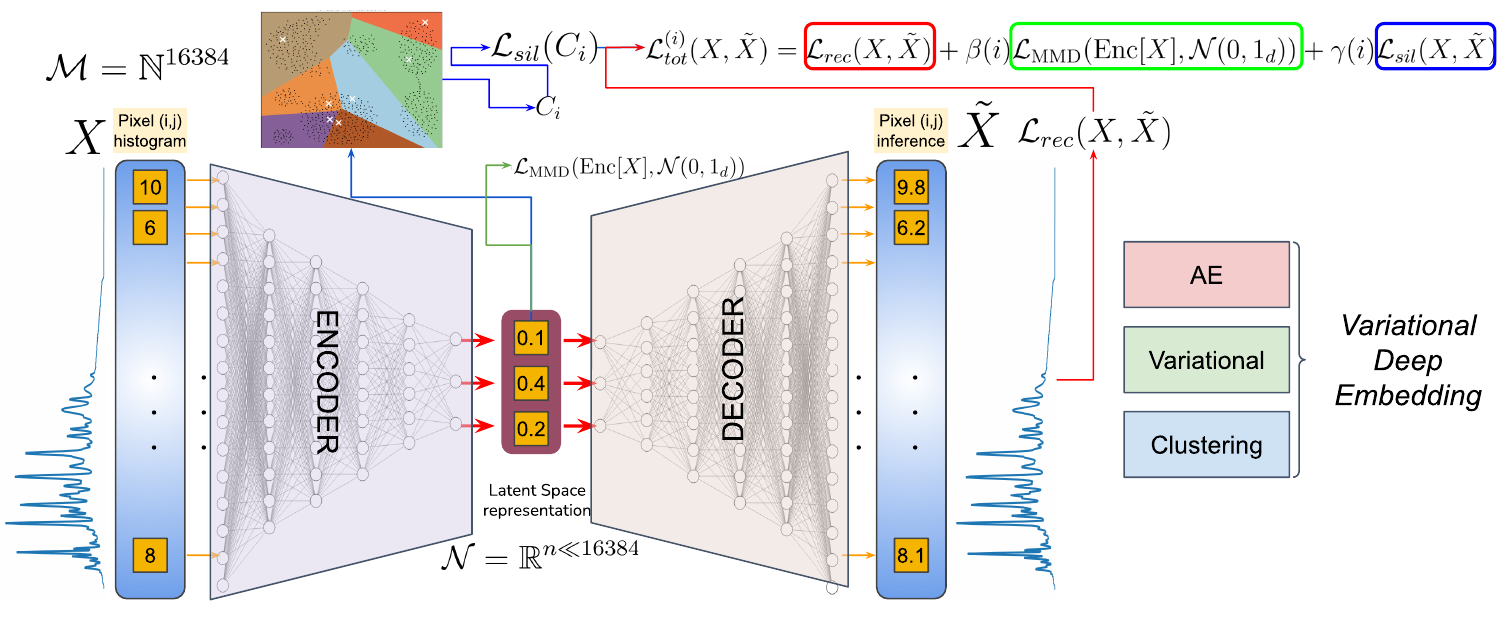}
    \caption{Visual representation of the Deep Variational Embedding model}
    \label{fig:DeepVAE_arch}
\end{figure}

The idea of this part of the pipeline is to train a Deep Variational Embedding model \cite{song2013auto, mrabah2020deep, yang2017kmeansfriendly, guo2017improved, bombini2024datacube, Dirks2022AutoEncoderNN, Higgins2016betaVAELB, Goodfellow-et-al-2016, Kingma_2019, BengioRepLernAE, prince2023understanding, jiang2017variational} to dimensionally reduce the signal via relevant feature extraction; The whole process is self-supervised, since only spectra are used during training. The whole architecture comprises:
\begin{enumerate}
    \item An Encoder $\enc : X \in \mathcal{M}\subseteq \mathbb{R}^{d=512} \mapsto \mu \in \mathcal{N}\subseteq \mathbb{R}^{n \ll 512}$, $\mu = \enc [X]$; 
    \item A clustering algorithm in latent space, $\ikmeans: \mu^{(i)} \mapsto C_{I=1, \ldots N}$; in this latent space, we can compute
        \begin{enumerate}
            \item  a self-supervised clustering loss, based on the \textit{silhuette score} \cite{silhouettescore1990};
                \begin{equation}
                    \mathcal{L}_{sil} (X, \tilde X) \equiv \frac{1 - \mathrm{silhouette}[C_{I=1, \ldots k}]}{2} \,, \quad C_{I=1, \ldots k} \leftarrow \ikmeans[\enc[X]] \,.
                \end{equation}
            \item a variational loss, inspired by the infoVAE model \cite{zhao2018infovae, Zhao2019InfoVAEBL}, which is the Maximum-Mean Discrepancy (MMD)  \cite{gretton2008kernel, li2015generative};
                \begin{equation}
                    \mathrm{MMD} ( q || p) = \mathbf{E}_{p(z), p(z')} [k(z, z')] - 2  \mathbf{E}_{q(z), p(z')} [k(z, z')] + \mathbf{E}_{q(z), q(z')} [k(z, z')] .
                \end{equation}
        \end{enumerate}
    \item A Decoder $\dec : \mu \in \mathcal{N}\subseteq \mathbb{R}^{n \ll 512} \mapsto  X \in \mathcal{M}\subseteq \mathbb{R}^{d=512} $, $\tilde X = \dec [\mu]$;
        \begin{enumerate}
            \item Compute a self-supervised reconstruction loss;
                \begin{equation}
                    \mathcal{L}_{rec} (X, \tilde X) = \frac{1}{N_{batch}} \sum_{a=1}^{N_{batch}} \frac{1}{d} \sum_{i=1}^d \left| X_{a, i} - \tilde X_{a, i} \right|^2 \,.
                \end{equation}
        \end{enumerate}
\end{enumerate}
Thus, the full loss we use, inspired by $\beta-$VAEs \cite{Higgins2016betaVAELB}, is 
\begin{equation}
    \mathcal{L}_{tot}^{(i)} (X, \tilde X) = \mathcal{L}_{rec} (X, \tilde X) + \beta(i) \mathcal{L}_{\mathrm{MMD}} (\enc[X], \mathcal{N}(0, \mathbbold{1}_d))    + \gamma (i) \mathcal{L}_{sil} (X, \tilde X),
\end{equation}
where $i$ is the $i-$th training epoch, and where $\mathcal{L}_{\mathrm{MMD}}  \equiv \mathrm{MMD} (\enc[X] || \mathcal{N}(0, \mathbbold{1}_d) )$.

This architecture is based on the Deep Clustering Network (DCN) architecture of \cite{yang2017kmeansfriendly} and the Variational Deep Embedding (VaDE) model of \cite{jiang2017variational}. 

The model hyperparameters were obtained using a grid-search method. The Encoder has 4+1 layer, while the decoder has 4 layers; the latent space has dimension $3$, the Multi-Layer Perceptrons (MLPs) are Self-normalising Neural Networks \cite{klambauer2017selfnormalizing}, and their sizes are $512\mapsto 256\mapsto  128\mapsto  64\mapsto  32\mapsto 3+3$ for the encoder, and $3 \mapsto  64 \mapsto 128 \mapsto  256 \mapsto 512$ for the decoder. Furthermore, $\gamma(i) = 0.01$ $\forall i$, while $\beta(i) = 0.01 \cdot \Theta(i - 30)$. 

A visual representation of the Deep Variational Embedding model is reported in \cref{fig:DeepVAE_arch}.

\subsection{Preparing the synthetic dataset - part II: embedded MA-XRF images}\label{subsec:datasetII}

\begin{figure}[t]
    \centering
    \subfloat[\centering Example of an Embedded MA-XRF image]{{\includegraphics[width=5cm]{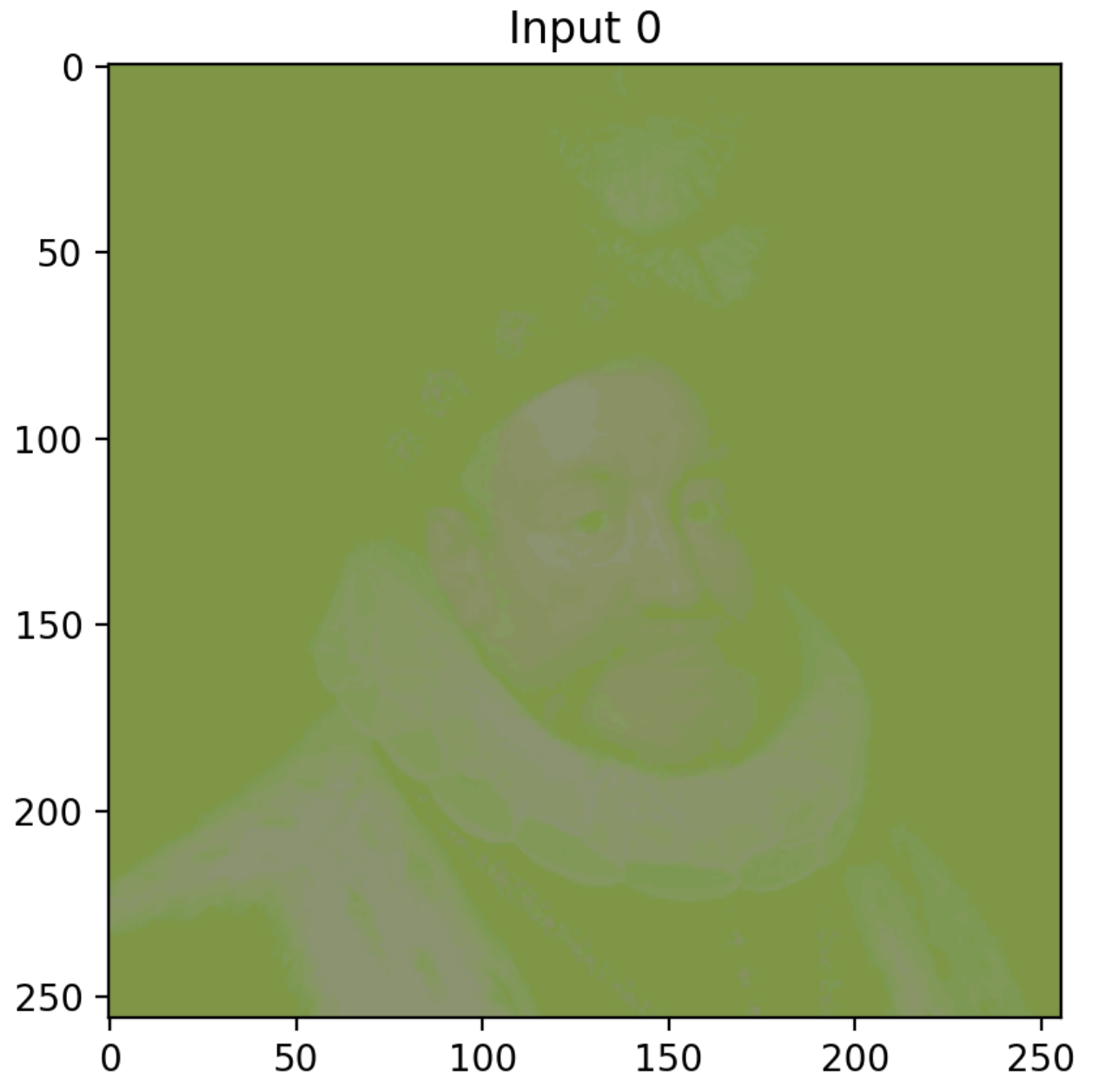} }}%
    \qquad
    \subfloat[\centering Relative RGB image]{{\includegraphics[width=5cm]{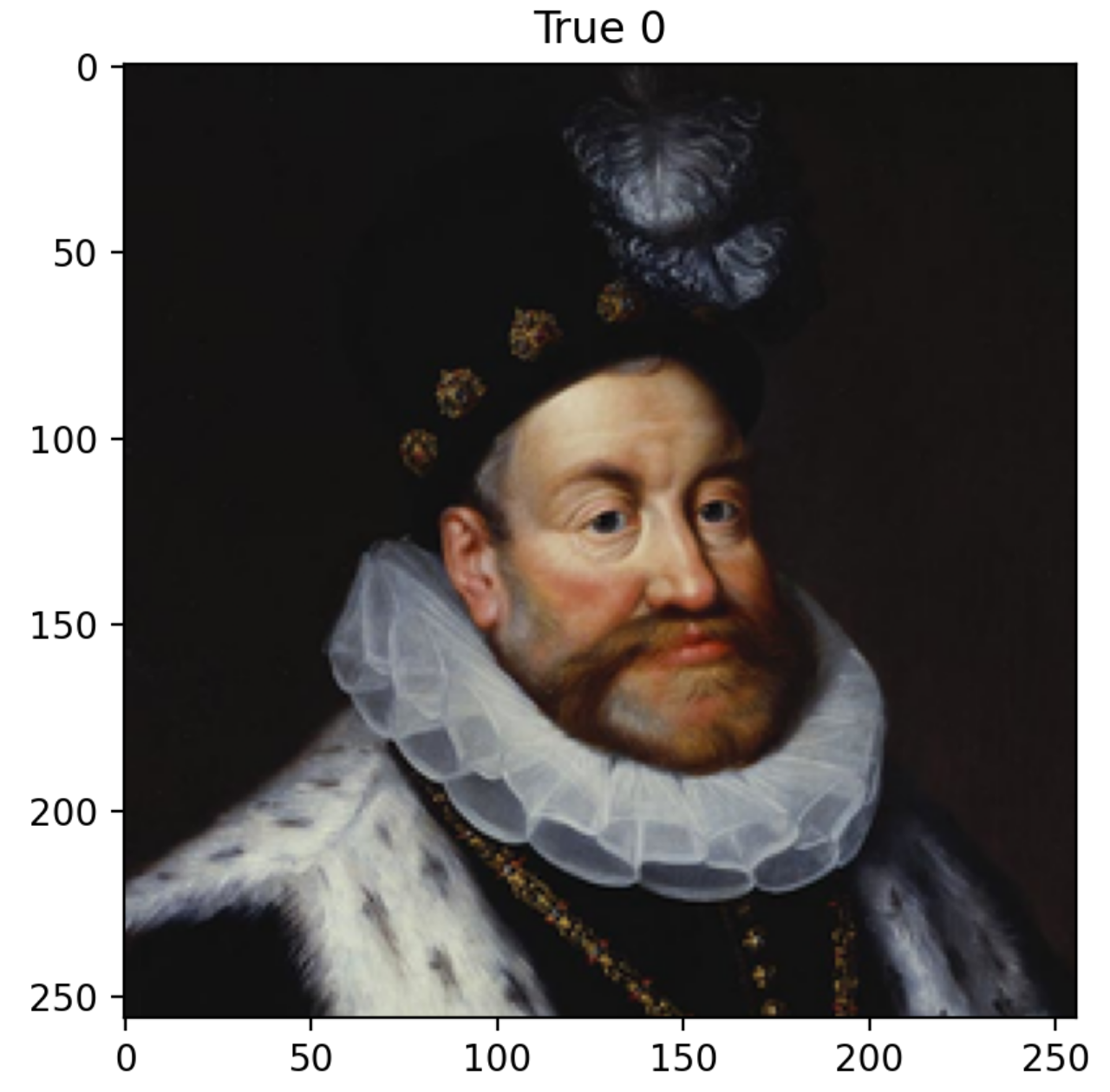} }}%
    \caption{Example of input-real output pair of the embedded dataset.}%
    \label{fig:example_data}%
\end{figure}

After having obtained a trained model following the procedure described in \cref{subsec:ae}, we can use it on the MA-XRF image dataset of \cref{subsec:datasetI}, and obtain a (disk occupation size reduced) dataset in embedded space, subbed embedded MA-XRF dataset.

This grants us a dataset of size (we used the \textsc{PyTorch} $[B, C, H, W]$ ordering)
\begin{itemize}
    \item Train set shape: \hspace{8.7mm}  \,$[208\,087, 3, 256, 256]$
    \item Validation set shape:    \, $[\;\;60\,466,  3, 256, 256]$
    \item Test set shape:  \hspace{11.7mm}   $[\;\;30\,210,  3, 256, 256]$
\end{itemize}

Please notice that the embedded MA-XRF images have three channels, but are \textit{not} RGB images; it was a mere coincidence that the embedded space have the same dimensions as the target RGB space. Nevertheless, the approach described here would still work even if the two spaces have different dimensions.

In \cref{fig:example_data} we report an example of a couple input-output $[X,Y]$. Notice that, due to the coincidence that the embedded space has 3 dimension, we can use false colours to represent the embedded MA-XRF.

\subsection{Training the virtual recolouring model}\label{subsec:vit}
We are now in the position to define a virtual recolouring model, whose goal is to map embedded MA-XRF images obtained in \cref{subsec:datasetII} to RGB images. 

\begin{figure}[ht]
    \centering
    \includegraphics[width=0.95\textwidth]{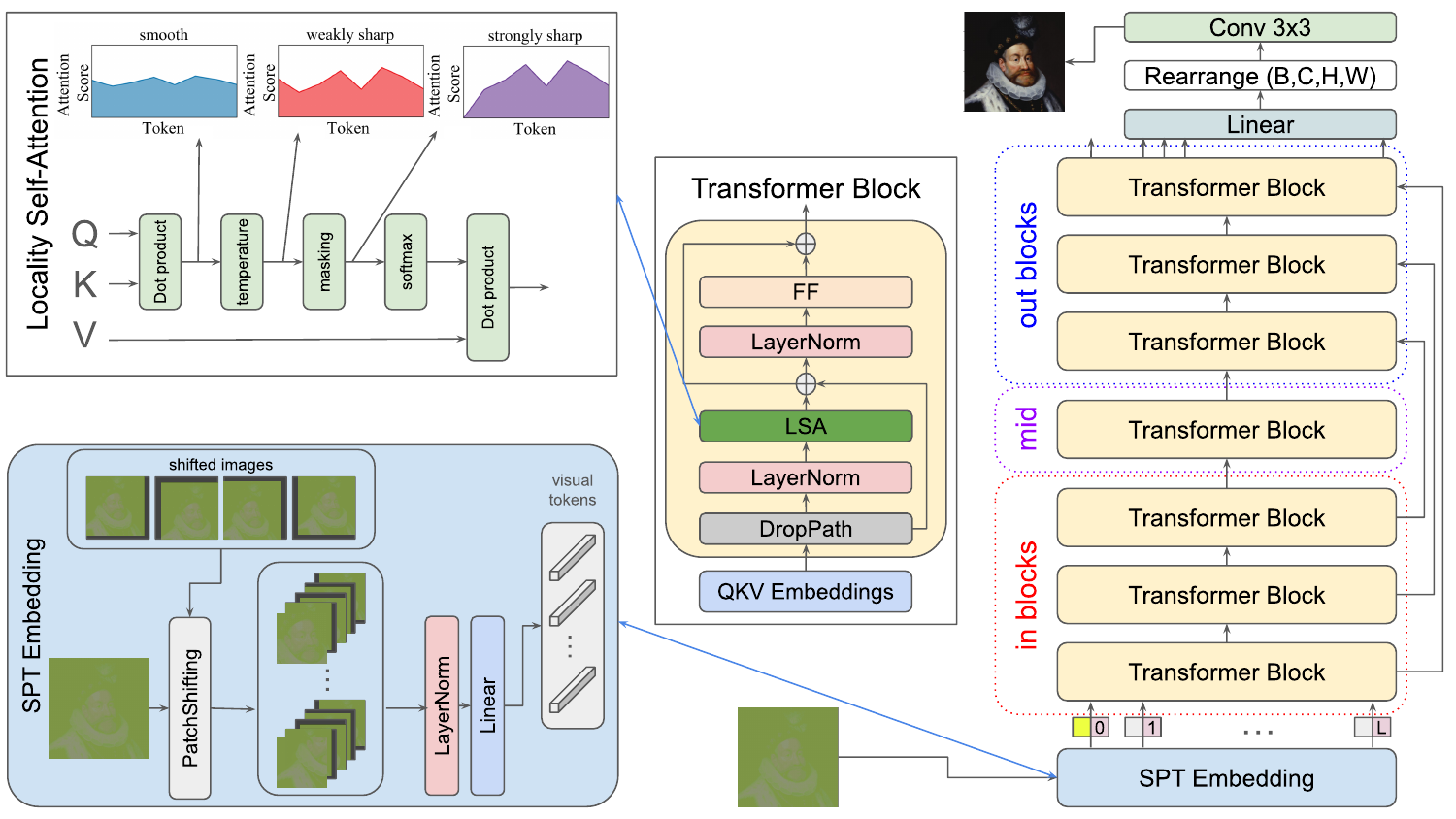}
    \caption{Graphical representation of the {SmallUViT} model.}
    \label{fig:enter-label}
\end{figure}

As a model, we use a Vision Transformer (ViT) \cite{dosovitskiy2021image} (for a nice survey, review, and introduction on Vision Transformers and their applications, see \cite{Han2020ASO, Li2023TransformerBasedVS, thisanke2023semantic, SHAMSHAD2023102802, drones7050287, touvron2021training}, and references therein). In particular, due to the limited size of the training dataset, we use a model based on the one presented in \cite{lee2021vision}; \ie, we use \textit{Shifted Patch Tokenization} (SPT) embedding, and \textit{Locality Self-Attention} (LSA), whose aim is to solve the lack of locality inductive bias, and enable ViT models to learn from scratch on reduced-size datasets.  

Furthermore, we employed long skip connections to form a U-ViT inner architecture \cite{bao2023worth}. Each transformer block in the backbone has thus a LSA multi-head part, and a Feed Forward (FF) part; the Feed Forward part is comprised of a sequential model of Linear Layer, Dropout layer \cite{JMLR:v15:srivastava14a}, GELU activation \cite{hendrycks2023gaussianerrorlinearunits}, followed again by a Linear layer and a dropout layer. 

After the U-shaped transformer layers, we have a Linear layer followed by a rearranging layer of the elaborated visual tokens (done using the \textsc{einops} package \cite{rogozhnikov2022einops}). The reconstructed tensor of shape $(B, C, H,W)$ is thus passed to a final convolutional layer with $3\times3$ kernel.

For simplicity, we will refer in thefollowing to this architecture as \textbf{SmallUViT} (U-Net based Vision Transformer for Small dataset tasks).

We used \textit{Stochastic Depth} during training \cite{huang2016deep, wightman2021resnet} (in an implementation inspired to the \textit{PyTorch Image Models } (timm)  library \cite{rw2019timm}, as done in the UViT paper \cite{bao2023worth}, following the recipe of the DeiT paper \cite{touvron2021trainingdataefficientimagetransformers}). We have also used data augmentations during training using the standard \textsc{torchvision} transformations \cite{TorchVision_maintainers_and_contributors_TorchVision_PyTorch_s_Computer_2016}. 

During training, the learning rate is changed dynamically via the \textsc{PyTorch} native \textsc{ReduceLROnPLateu} function by monitoring the selected metric, which is the \textit{Multi-Scale Structural Similarity Index} (MS-SSIM) \cite{MSSSIM1292216, DBLP:journals/corr/ZhaoGFK15} in the Lightning AI \textsc{torchmetrics} implementation \cite{torchmetrics_2022}.

We arrived at the formulation of this architecture by trial-and-error on multiple architectures, starting from the plain ViT \cite{dosovitskiy2021image}, the UViT \cite{bao2023worth}, the SWIN transformer \cite{liu2021swintransformerhierarchicalvision}, and the SEgmentation TRansformer (SETR) with progressive up-sampling design (SETR-PUP) \cite{zheng2021rethinkingsemanticsegmentationsequencetosequence}.  We trained various implementation of such architecture (and mixture of them) on a reduced dataset (sampled from the full one), in a plain grid-search approach, finding the most performing one at (almost) fixed number of parameters, \ie the aforementioned SmallUViT\footnote{In this exploratory phase, we started by training convolutional-based networks which, at fixed model size, had worse performances on the loss and metric used.}.

\subsubsection{A note on colour distance and training loss: sRGB} \phantom{.}

To train the SmallUViT model we have used an ad hoc distance for the loss, to take in account perceptual differences in colors; due to its simple implementation, we used the CompuPhase \textit{redmean} sRGB distance approximation \cite{sRGB_coode}:
\begin{align}
    \delta(C_1, C_2) &= \sqrt{ (2+r) \Delta r^2 + 4 \Delta g^2 + (3 - r)\Delta b^2 } \,, \quad r \equiv \frac{r_1 + r_2}{2} \,,
\end{align}
where $C_i = [r_i, g_i, b_i]$, $\Delta r \equiv r_2 - r_1$, and similarly for the other channels. 

Thus the loss is
\begin{equation}
    \mathcal{L}_{\mathrm{sRGB}} (y, \hat{y}) = \frac{1}{H\cdot W} \sum_{h,w} \frac{1}{N_b} \sum_{b=1}^{N_b} \delta (y_{b, c, h, w}, \hat y_{b, c, h, w} )\,,
\end{equation}
where $y_{b,c,h,w} = \mathrm{SmallUViT}[x_{b,c',h,w}]$ is the model prediction, while $\hat y_{b,c,h,w}$ is the ground truth.

\section{Results: from MA-XRF to RGB}\label{sec:results}

Among the possible hyperparameter set selection at fixed VRAM occupation size we could handle with our set-up at this stage\footnote{We Trained the model on a NVIDIA Tesla T4 GPU with 16 Gb vRAM, located on a node with 128 Gb RAM, 2x Intel Xenon Gold CPU.}, we selected the best model architecture via a plain grid search. 

The selected model has 3 485 076 trainable parameters, obtained by 3+1+3 Transformer layers in the in/mid/out U-ViT blocks, a $16\times16$ patch size, 9 heads per MultiHead block, 192 embed dimension size, 2 MLP-to-Dim factor, and a head dimension of 32. The dropout rate and the stochastic depth rate are both set to $0.1$. 

We trained the SmallUViT model for 100 epochs, using the schemes described in \cref{subsec:vit}; 

\begin{figure}
    \centering
    \includegraphics[width = 0.8 \textwidth]{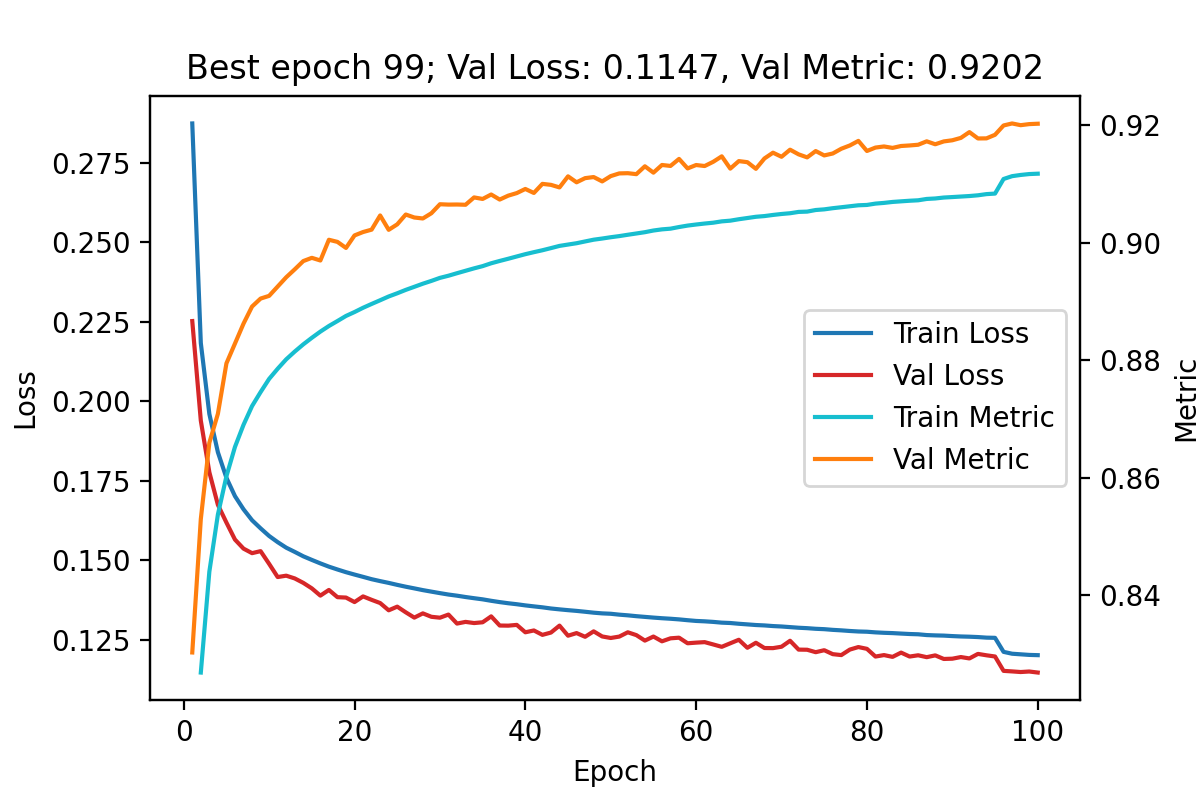}
    \caption{Training history Plot}
    \label{fig:training_history_plot}
\end{figure}

The result of the training is reported in \cref{fig:training_history_plot}; the discrepancy between validation and train loss/metric may be traced back to the presence of Dropout and Stochastic Depth during training. 

Applied on the $30\,210$ images in the test set, we get a sRGB loss of $\mathcal{L}_{\text{test}} = 0.1148$, and an MS-SSIM metric value of $\mathcal{M}_{\text{test}} = 0.9196$.

\subsection{Application on some test set cases}

\begin{figure}[ht]
    \centering
    \includegraphics[width=0.95\textwidth]{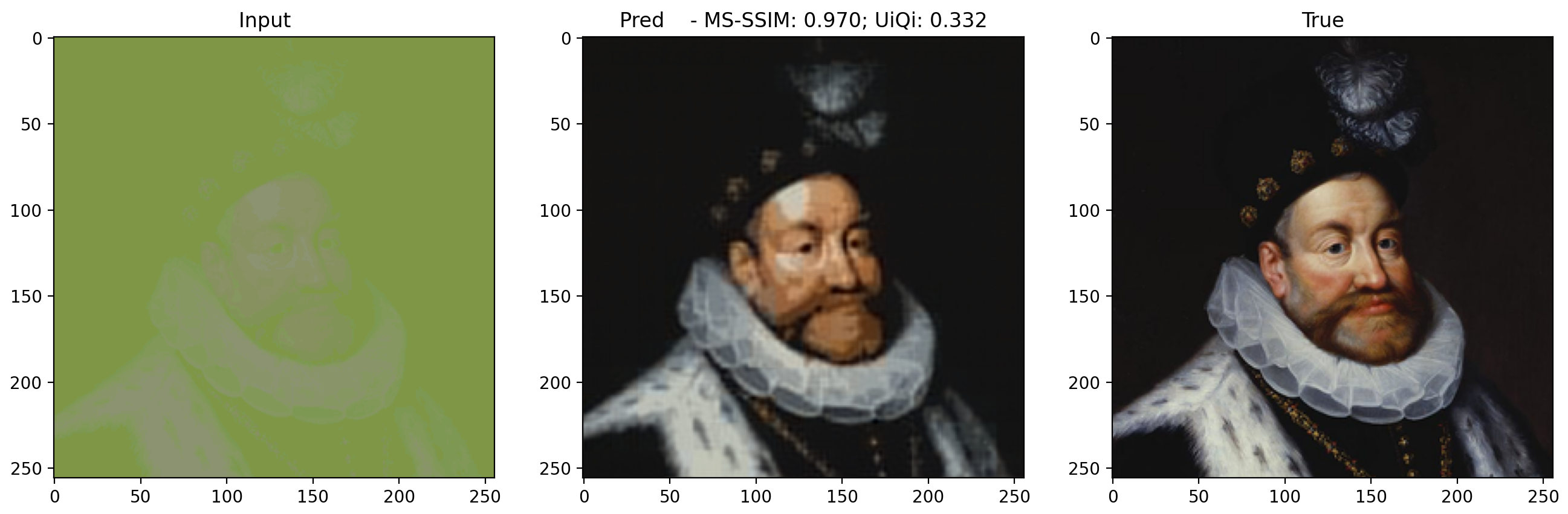}
    \caption{Example number 0}
    \label{fig:SmallUViT_NEW_PALETTE_inference_0}
\end{figure}

\begin{figure}[ht]
    \centering
    \includegraphics[width=0.95\textwidth]{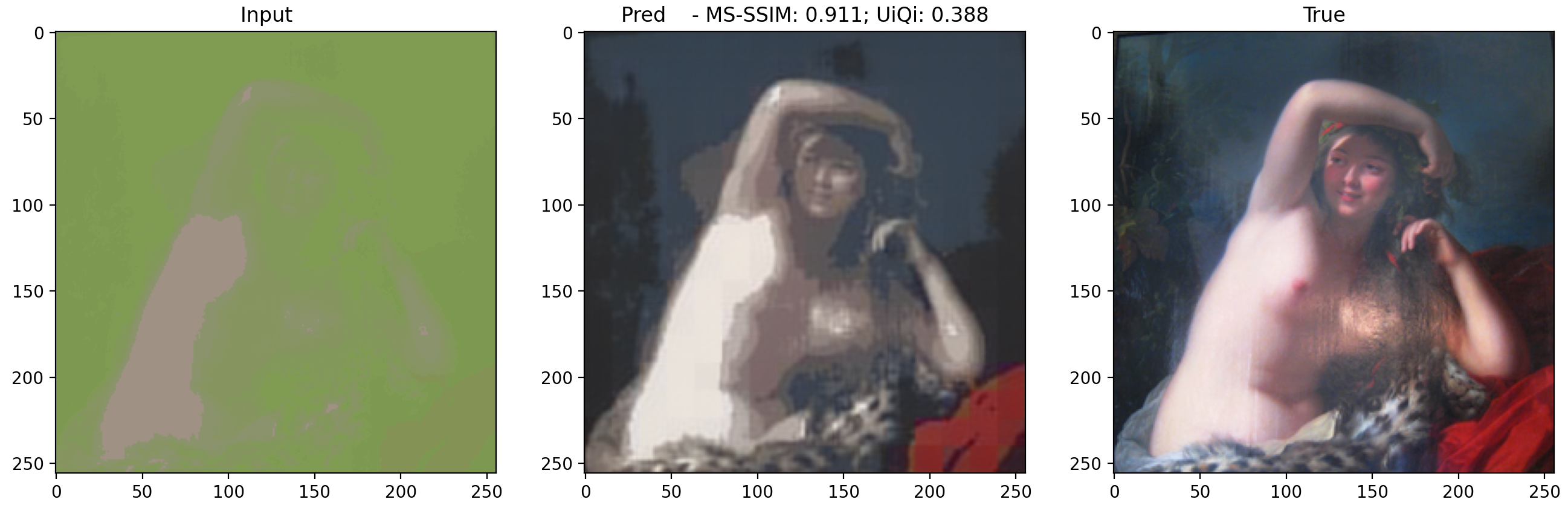}
    \caption{Example number 1}
    \label{fig:SmallUViT_NEW_PALETTE_inference_2}
\end{figure}

\begin{figure}[ht]
    \centering
    \includegraphics[width=0.95\textwidth]{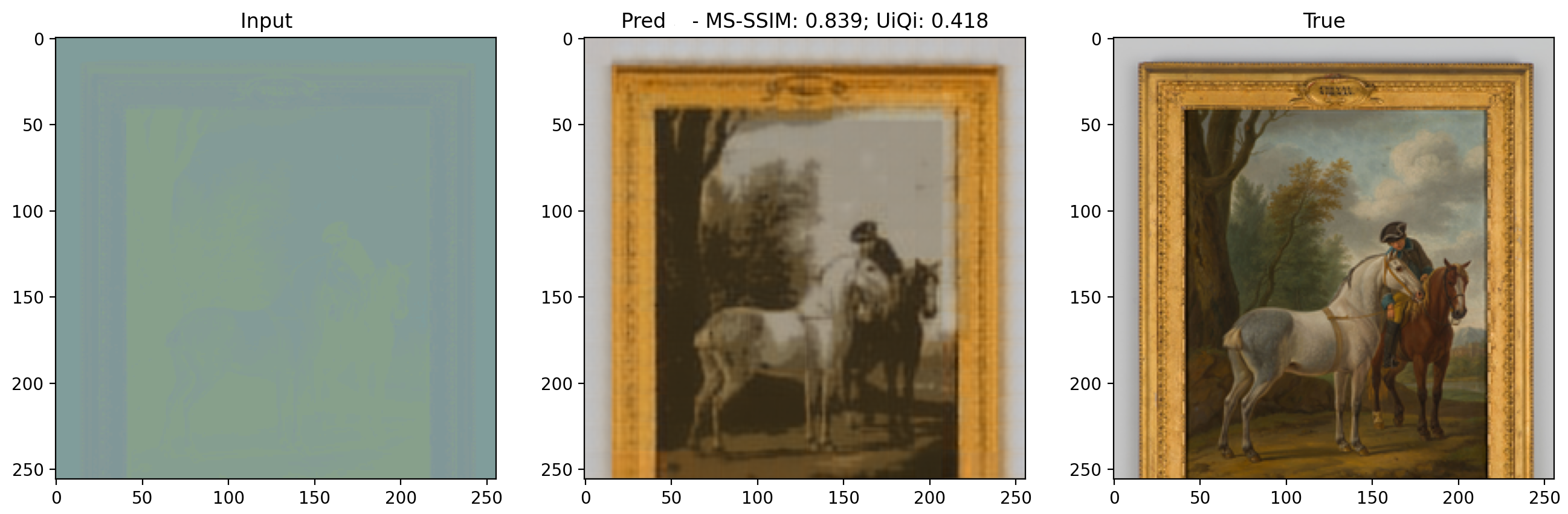}
    \caption{Example number 2}
    \label{fig:SmallUViT_NEW_PALETTE_inference_6}
\end{figure}

\begin{figure}[ht]
    \centering
    \includegraphics[width=0.95\textwidth]{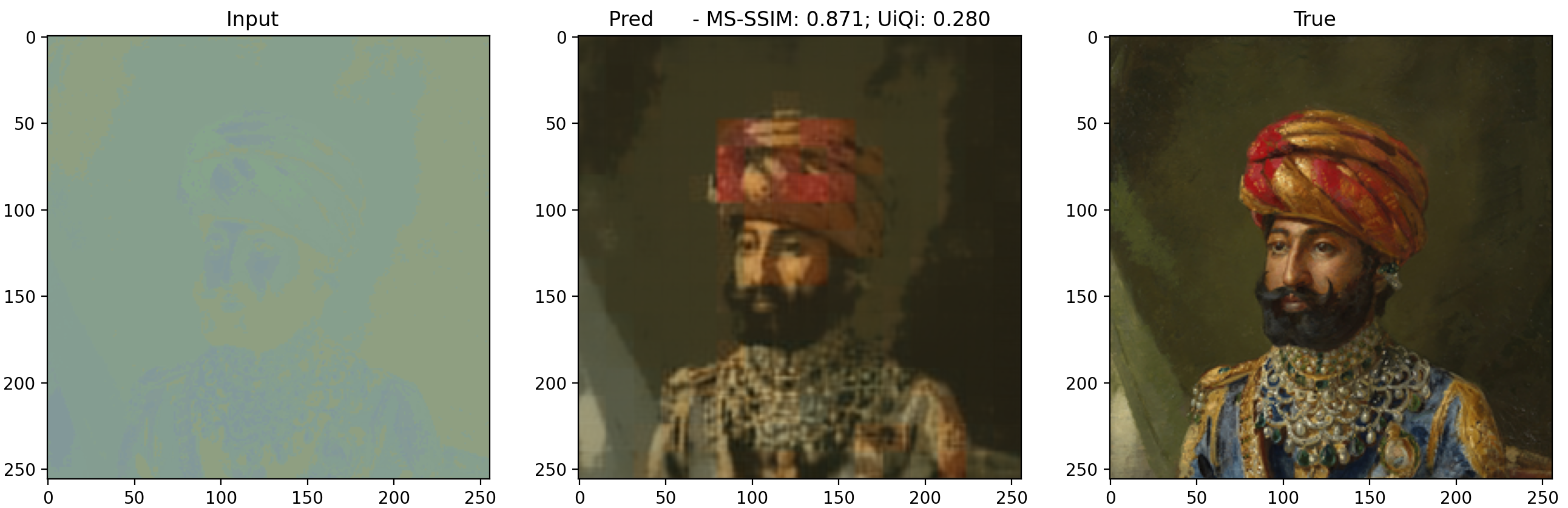}
    \caption{Example number 3}
    \label{fig:SmallUViT_NEW_PALETTE_inference_15}
\end{figure}

We now report few examples extracted from the test dataset. For each example, we show a triplet of figures: the leftmost one, is the embedded   synthetic MA-XRF image; the middle one, is the output of the SmallUViT; the rightmost one, is the true RGB image. On top of the image in the middle are reported two metric scores, computed between the prediction and the target: the Multi-Scale Similarity Index, and the \textit{Universal Image Quality Index} (UiQi) \cite{UiQi995823}. The four examples presented in \cref{fig:SmallUViT_NEW_PALETTE_inference_0,fig:SmallUViT_NEW_PALETTE_inference_2,fig:SmallUViT_NEW_PALETTE_inference_6,fig:SmallUViT_NEW_PALETTE_inference_15} are selected because they present relevant feature of the inference of the model (pros and cons); they are not the ones with the higher MS-SSIM nor UiQi scores in the test dataset. Furthermore, we choose to show the comparison of the inferred RGB with the actual (resized) image, and not with its $\ikmeans$ clustered version employed in the synthetic dataset generating algorithm (we refer to \cref{alg:GenerateMA-XRF}).

From the presented images, we already seen pros and cons of the trained model. Even if the model was limited, the pipeline cumbersome, the task complex and the computing power limited, we were able to extrapolate a set of RGB images out of (embedded, synthetic) MA-XRF data cubes. This is quite comforting. 

Nevertheless, even if the MS-SSIM score is quite high (around $0.9$), and, overall, the network was capable of assigning colors to semantic regions of the images, we may spot few issues, as partially signalled by the low UiQi scores. 

\begin{figure}[t]
    \subfloat[Clustered RGB 0]{%
        \includegraphics[width=.45\linewidth]{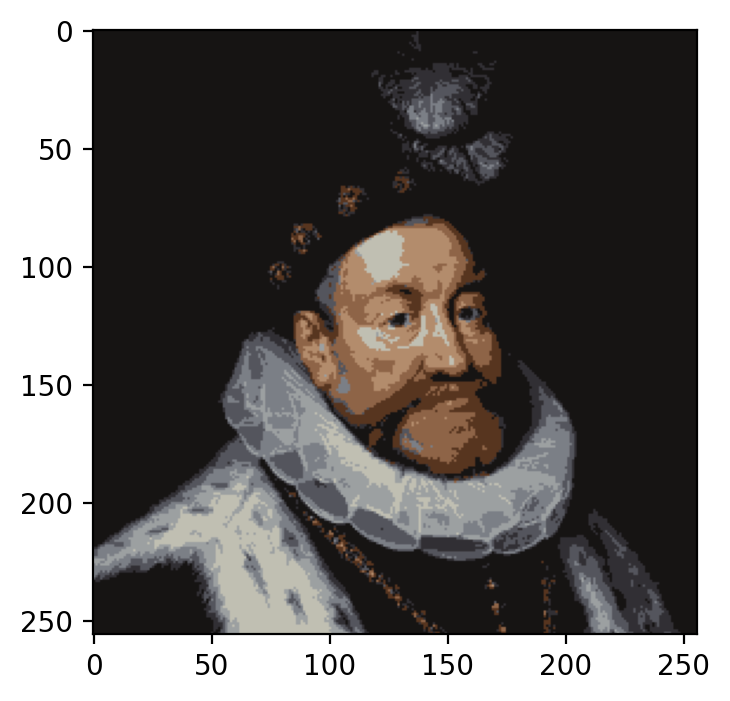}
        \label{subfig:clustered_rgb_0}%
    }\hfill
    \subfloat[Clustered RGB 1]{%
        \includegraphics[width=.45\linewidth]{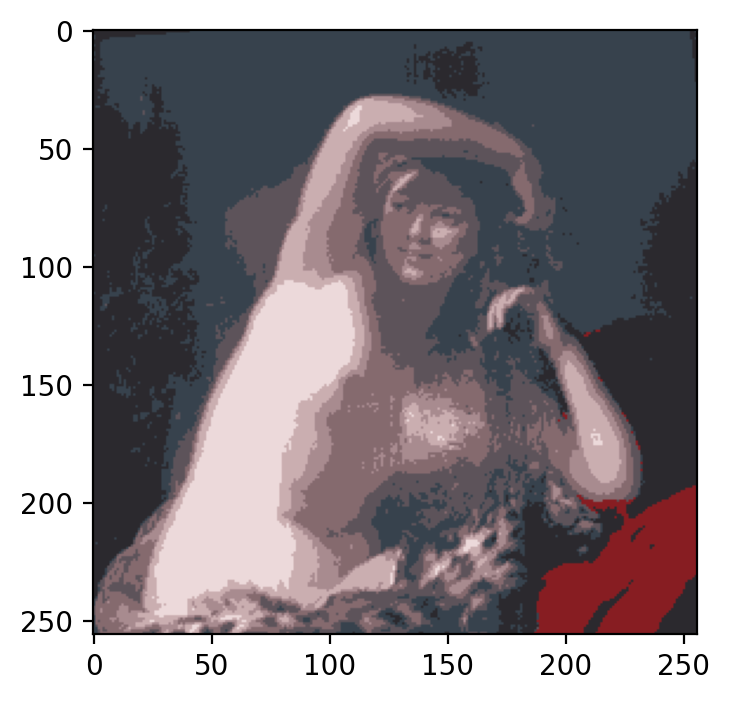}%
        \label{subfig:clustered_rgb_1}%
    }\\
    \subfloat[Clustered RGB 2]{%
        \includegraphics[width=.45\linewidth]{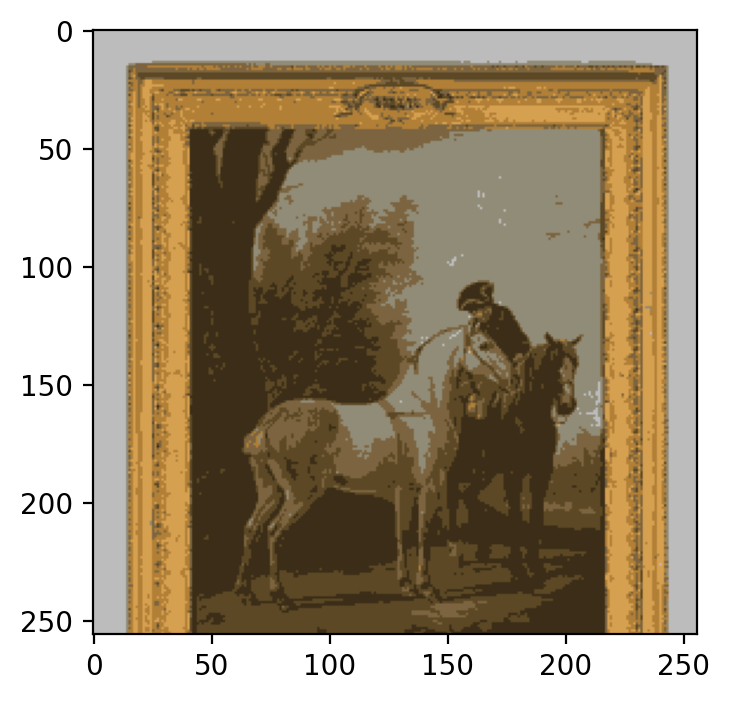}%
        \label{subfig:clustered_rgb_2}%
    }\hfill
    \subfloat[Clustered RGB 2]{%
        \includegraphics[width=.45\linewidth]{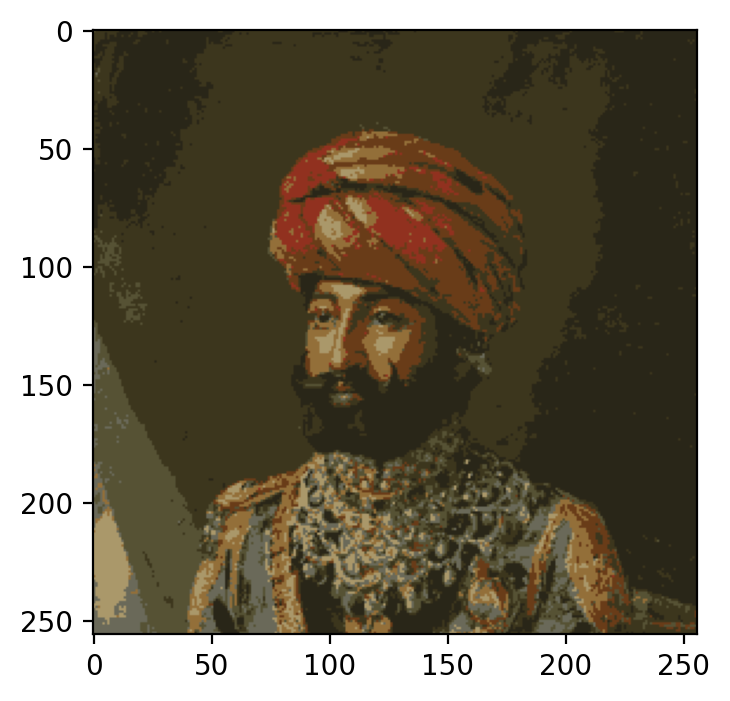}%
        \label{subfig:clustered_rgb_3}%
    }
    \caption{Clustered RGB test images.}
    \label{fig:clustered_rgbs}
\end{figure}

Interesting examples are \cref{fig:SmallUViT_NEW_PALETTE_inference_0}, \cref{fig:SmallUViT_NEW_PALETTE_inference_2}; the network was able of identifying the dark colours (the blueish black and the dark blue of the background, as well as the black of the jacket), the light colours (the white and the light blue of the man's Ruff), and, quite surprisingly, the red, which is given (in the example) by the embedding of signals formed (in part) by a red ochre, which is, in composition, quite similar to the Gold Ochre\footnote{At least, in the spectra contained in the database we use. See \url{https://infraart.inoe.ro/sample/1578/show/} and \url{https://infraart.inoe.ro/sample/130/show/} for yellow and red ochre, respectively. From \cite{infraart10.1145/3593427}.}.  Yet, the incarnate has been not properly coloured, at least in \cref{fig:SmallUViT_NEW_PALETTE_inference_2} (instead, for the peculiar visual aspect of \cref{fig:SmallUViT_NEW_PALETTE_inference_0}, the $\ikmeans$ clustering performed on the RGB has to be blamed, see \cref{subfig:clustered_rgb_0}). The poor performance on such incarnate may also be attributed to the complex pigment selection during this image generation - fact that can be hinted from the false rgb representation of the embedded synthetic MA-XRF. 

In \cref{fig:SmallUViT_NEW_PALETTE_inference_6}, the model correctly coloured the frame's yellow (a difficult pigment), the leaves' greens are quite correctly identified and, apart from a slight colour deviation, so are the ground and horses colours. Yet, no clouds appears in the sky. This is due to the aforementioned RGB clustering (see \cref{subfig:clustered_rgb_2}). 

Finally, \cref{fig:SmallUViT_NEW_PALETTE_inference_15} presents an interesting artefact. While the background greens, the incarnate, the beard, and, partly, the uniform colours have been partly identified, the turban red posed a difficult task to the network, and it creates a patching artefact. 

\section{Conclusions}\label{sec:concl}

In this contribution we report the design and a first test of a pipeline aimed at virtual recolour MA-XRF data cubes. This task is intended to furnish restoration scientist and heritage scientist a visual aid to help them elaborate the raw data obtained from nuclear imaging performed on pictorial artworks, especially in those situation where the study object suffers from loss of visual readability. 

The goal of the pipeline design is to perform the task starting from synthetic data, to overcome the lack of availability of data. Furthermore, the pipeline has an intermediate step to  (a) address the high per-MA-XRF image disk size, and (b) extract relevant spectral features, regardless of experimental/environmental conditions. 

Thus, the designed pipeline comprises two trained models: 
\begin{itemize}
    \item A Deep Embedding model working on single spectra;
    \item A Vision Transformer to elaborate embedded.
\end{itemize}

We have shown that, a part from some difficulties inherited by the synthetic dataset generation algorithm, we have a pipeline capable of performing the end-to-end task. 

\subsection{Next steps}

After having motivated the pipeline design and architecture, the next step is apply it on a recolouring use case task. To do so, the crucial part is devising an appropriate \textit{domain adaptation} technique to handle the domain shift between the synthetic dataset and the real MA-XRF data. The fact that the pipeline relies on two networks allows us to perform the domain adaption unsupervisingly on one training only, \ie add the domain adaptation (\eg, as an adversarial step as shown in \cite{sun2023domainadaptationadversarialtraining}), while also giving us the opportunity the tackle each issues arising from the network disjointly and in parallel. 

The first and foremost issue it is possible to trace back, is the limited size of the dataset. We plan to enlarge it by removing the restriction of scraped images from WikiData to be either paintings or frescoes. This, of course, alters the domain  of the dataset, including scenes not usually represented in the domain of  application of the network (\ie, pictorial artworks). Nevertheless, the goal is to split the train into two parts:
\begin{enumerate}
    \item A first self-supervised step, to force the ViT to learn the semantic aspects of images (using the enlarged dataset);
    \item A second step, where the pretrained network should learn how to assign colours (using the pictorial artwork dataset). 
\end{enumerate}

Furthermore, in \cref{sec:results} we have reported few limitations shown by the network; mainly, its inability to reconstruct visual appearances lost during the rgb clustering step in the synthetic dataset generation algorithm. One way to tackle this issue would be to add a \textit{generative} part to the model, in order to infer something not properly encoded in the model's input. To do so, we plan to implement a \textsc{Pix2Pix} approach \cite{isola2018imagetoimagetranslationconditionaladversarial}, using SmallUViT as backbone (for a review on GAN approaches with ViT, see \cite{dubey2023transformerbasedgenerativeadversarialnetworks} and references therein).

\section{Code and Data Availability}\label{sec:data-and-code}

The code used in this project can be found at the ICSC Spoke 2 repository \url{https://baltig.infn.it/chnet/fast-extended-vision-smalluvit}.

\section*{Acknowledgements}

To conduct the work presented in this contribution, we resorted to the cloud computing Software-as-a-Service infrastructure made available by the ML\_INFN initiative (“Machine Learning at INFN”) \cite{ML_INFN_paper}, aim to foster Machine Learning activities at the Italian National Institute for Nuclear Physics (INFN), as well as some servers made available by INFN-CHNet \cite{app11083462}, the network devoted to the application of nuclear techniques to Cultural Heritage.

We would also like to thank the authors of the site \url{https://infraart.inoe.ro/} \cite{infraart10.1145/3593427}, and WikiData \url{https://www.wikidata.org} and WikiMedia commons \url{https://commons.wikimedia.org}.

\subsection*{Funding}
This work is supported by ICSC – Centro Nazionale di Ricerca in High Performance Computing, Big Data and Quantum Computing, funded by European Union – NextGenerationEU, by the European Commission within the Framework Programme Horizon 2020 with the project ``4CH - Competence Centre for the Conservation of Cultural Heritage"  (GA n.101004468 – 4CH) and by the project AIRES–CH - Artificial Intelligence for digital REStoration of Cultural Heritage jointly funded by the Tuscany Region (Progetto Giovani Sì) and INFN.

The work of {AB} was funded by Progetto ICSC - Spoke 2 - Codice CN00000013 - CUP I53C21000340006 - Missione 4 Istruzione e ricerca - Componente 2 Dalla ricerca all'impresa – Investimento 1.4.

The work of {FGAB} was funded by the research grant titled ``Artificial Intelligence and Big Data'' and funded by the AIRES-CH Project cod.~291514 (CUP I95F21001120008).

\clearpage
\appendix

%
%
\clearpage 
\section*{References}
\bibliographystyle{iopart-num}
\bibliography{main}
\end{document}